\documentclass[lettersize,journal]{IEEEtran}
\usepackage{amsmath,amsfonts}
\usepackage{algorithmic}
\usepackage{algorithm}
\usepackage{array}
\usepackage[caption=false,font=normalsize,labelfont=sf,textfont=sf]{subfig}
\usepackage{textcomp}
\usepackage{stfloats}
\usepackage{url}
\usepackage{verbatim}
\usepackage{graphicx}
\usepackage{cite}
\usepackage{booktabs}
\usepackage{multirow}
\usepackage{scalerel}

\usepackage{color}
\usepackage[dvipsnames]{xcolor}

\usepackage{expl3}
\ExplSyntaxOn
\newcommand\latinabbrev[1]{
  \peek_meaning:NTF . {
    #1\@}%
  { \peek_catcode:NTF a {
      #1.\@ }%
    {#1.\@}}}
\ExplSyntaxOff

\def\ie{\latinabbrev{\emph{i.e}}}
\def\vs{\latinabbrev{\emph{v.s.}}}

\begin{document}

\title{Fast Adversarial Training with Noise Augmentation: \\ A Unified Perspective on RandStart and GradAlign}

\author{Axi Niu, Kang Zhang, Chaoning Zhang*, Chenshuang Zhang, In So Kweon, ~\IEEEmembership{Member, ~IEEE},\\ Chang D. Yoo~\IEEEmembership{Senior Member,~IEEE}, and Yanning Zhang,~\IEEEmembership{Senior Member,~IEEE}
\thanks{This work was funded in part by the Project of the National Natural Science Foundation of China under Grant 61901384 and 61871328, Natural Science Basic Research Program of Shaanxi under Grant 2021JCW-03, as well as the Joint Funds of the National Natural Science Foundation of China under Grant U19B2037.).

Axi Niu and Yanning Zhang are with the School of Computer Science, Northwestern Polytechnical University, Xi’an, 710072, China (email: nax@mail.nwpu.edu.cn, ynzhang@nwpu.edu.cn )

Kang Zhang, Chaoning zhang, Chenshuang Zhang, Chang D. Yoo and In So Kweon are with the School of Electrical Engineering, Korea Advanced Institute of Science and Technology, Daejeon, Republic of Korea (email:zhangkang@kaist.ac.kr, chaoningzhang1990@gmail.com,  zcs15@kaist.ac.kr, cd\_yoo@kaist.ac.kr, iskweon77@kaist.ac.kr)

* Corresponding author: Chaoning Zhang
}}

\markboth{Journal of \LaTeX\ Class Files,~Vol.~14, No.~8, August~2021}%
{Shell \MakeLowercase{\textit{et al}}: A Sample Article Using IEEEtran.cls for IEEE Journals}

\maketitle

\begin{abstract}
PGD-based and FGSM-based are two popular adversarial training (AT) approaches for obtaining adversarially robust models. Compared with PGD-based AT, FGSM-based one is significantly faster but fails with catastrophic overfitting (CO). For mitigating CO in such Fast AT, there are two popular existing strategies: random start (RandStart) and Gradient Alignment (GradAlign). The former works only for a relatively small perturbation $8/255$ with the $l_\infty$ constraint, and GradAlign improves it by extending the perturbation size to $16/255$ (with the $l_\infty$ constraint) but at the cost of being 3 to 4 times slower. 
How to avoid CO in Fast AT for a large perturbation size but without increasing the computation overhead remains as an unsolved issue, for which our work provides a frustratingly simple (yet effective) solution. Specifically, our solution lies in just noise augmentation (NoiseAug) which is a non-trivial byproduct of simplifying GradAlign. By simplifying GradAlign we have two findings: (i) aligning logit instead of gradient in GradAlign requires half the training time but achieves higher performance than GradAlign; (ii) the alignment operation can also be removed by only keeping noise augmentation (NoiseAug). Simplified from GradAlign, our NoiseAug has a surprising resemblance with RandStart except that we inject noise on the image instead of perturbation. 
To understand why injecting noise to input prevents CO, we verify that this is caused not by data augmentation effect (inject noise on image) but by improved local linearity. We provide an intuitive explanation for why NoiseAug improves local linearity without explicit regularization. Extensive results demonstrate that our NoiseAug achieves SOTA results in FGSM AT. The code will be released after accepted.
\end{abstract}


\section{Introduction}
\IEEEPARstart{D}{eep} neural networks are often vulnerable to adversarial examples (AEs) where the quasi-invisible adversarial perturbation causes misclassification~\cite{szegedy2013intriguing,amini2020towards,zhang2020interpreting,xu2021self,ye2021annealing}. Early attempts to improve adversarial robustness include various image processing techniques and detection techniques~\cite{yi2020improving,zhang2021joint,agarwal2021damad,du2022multiview,liu2021model}, however, most of them are found to give a false sense of robustness~\cite{carlini2017adversarial, athalye2018obfuscated,croce2020reliable,che2021adversarial}. 
In recent years, there is an emerging consensus that adversarial training (AT) and its variants are the most effective approaches for guaranteeing robustness against various strong white-box attacks~\cite{zhang2018face,zhang2021defense,liu2021training,wu2020augmented,gallego2020incremental,ma2019deep}. In contrast to standard training that directly solves a loss minimization problem, AT needs to first generate adversarial examples by solving an inner maximization problem. Seeking optimal adversarial examples, however, is an NP-hard problem~\cite{katz2017reluplex,weng2018towards,jia2022boosting,xu2022infoat,liang2021exploring}, for which AT adopts approximation methods. 

One widely adopted approximation method for the inner maximization problem is the PGD attack~\cite{madry2017towards} which generates adversarial examples by performing multi-step gradient update. AT with PGD attack solving the inner maximization problem is termed PGD AT which is widely believed to guarantee robustness to PGD attack as well as more recent AutoAttack~\cite{croce2020reliable}. This belief is corroborated by both theoretical verification for small models~\cite{carlini2017provably,tjeng2017evaluating,che2020new} and empirically unbroken robustness on public challenges for large models~\cite{madry2017towards,croce2020reliable}. Evaluating numerous recently proposed AT methods,~\cite{croce2020reliable,pang2020bag} have concluded that the claimed substantial robustness improvement in most of them is overestimated, rendering simple PGD AT still a mainstream approach. Despite its effectiveness, a major line of works attempt to discard it with the goal to make AT more computationally efficient. 

A straightforward way to improve efficiency of AT is to replace multi-step PGD attack with single-step FGSM attack~\cite{goodfellow2014explaining} for the inner maximization problem. Intuitively, if the quality of FGSM attack is sufficiently close to that of PGD attack in AT, FGSM AT is expected to also yield robust models as PGD AT. Given the linear nature of FGSM attack, one way to guarantee its quality is to adopt a smaller step size~\cite{kim2020understanding}. However, such a strategy only produces robustness to a relatively small perturbation.
Naively increasing the perturbation step size of FGSM attack will cause a failure mode termed catastrophic overfitting (CO), where the robustness suddenly and significantly drops in later stage of FGSM AT. To this end,
Wong et al proposed random start (RandStart)~\cite{wong2020fast}, \ie\ 
initializing the perturbation with random noise to make it compatible with a relatively larger perturbation size (like 8/255 with the $l_\infty$ constraint. RandStart still fails when the step size is larger than 10/255~\cite{wong2020fast}, and later~\cite{andriushchenko2020understanding} proposed a strategy termed Gradient Alignment (GradAlign) which shows success of FGSM AT for a large perturbation step size (like 16/255). As recognized in~\cite{andriushchenko2020understanding}, GradAlign has a major disadvantage of being 3 to 4 times slower than its counterpart with RandStart. This motivates us to consider the following challenge in FGSM AT as the main theme of this paper:
\\
\\
\textbf{Is there a method (a) as effective as GradAlign for a large perturbation size and (b) as efficient as RandStart without causing additional computation overhead?}
\\

To our best knowledge, GradAlign~\cite{andriushchenko2020understanding} is the only solution in the literature to show success of FGSM AT with a large perturbation size like $16/255$. Towards understanding what component(s) in GradAlign makes its success, we investigate whether a simplified GradAlign can still avoids CO. Our investigation leads to two important findings: (i)
aligning logit instead of gradient requires half the training time but achieves comparable (slightly superior) performance as GradAlign; (ii) the alignment operation can also be removed by only keeping noise augmentation (NoiseAug) to just add noise on input. Our findings suggest that the essential component in GradAlign for its success can be roughly reduced to its NoiseAug. This insight yields a simple yet effective/efficient solution to simultaneously satisfy the above criteria (a) and (b).

\begin{figure*}[ht]
\begin{center}
\includegraphics[width=2\columnwidth]{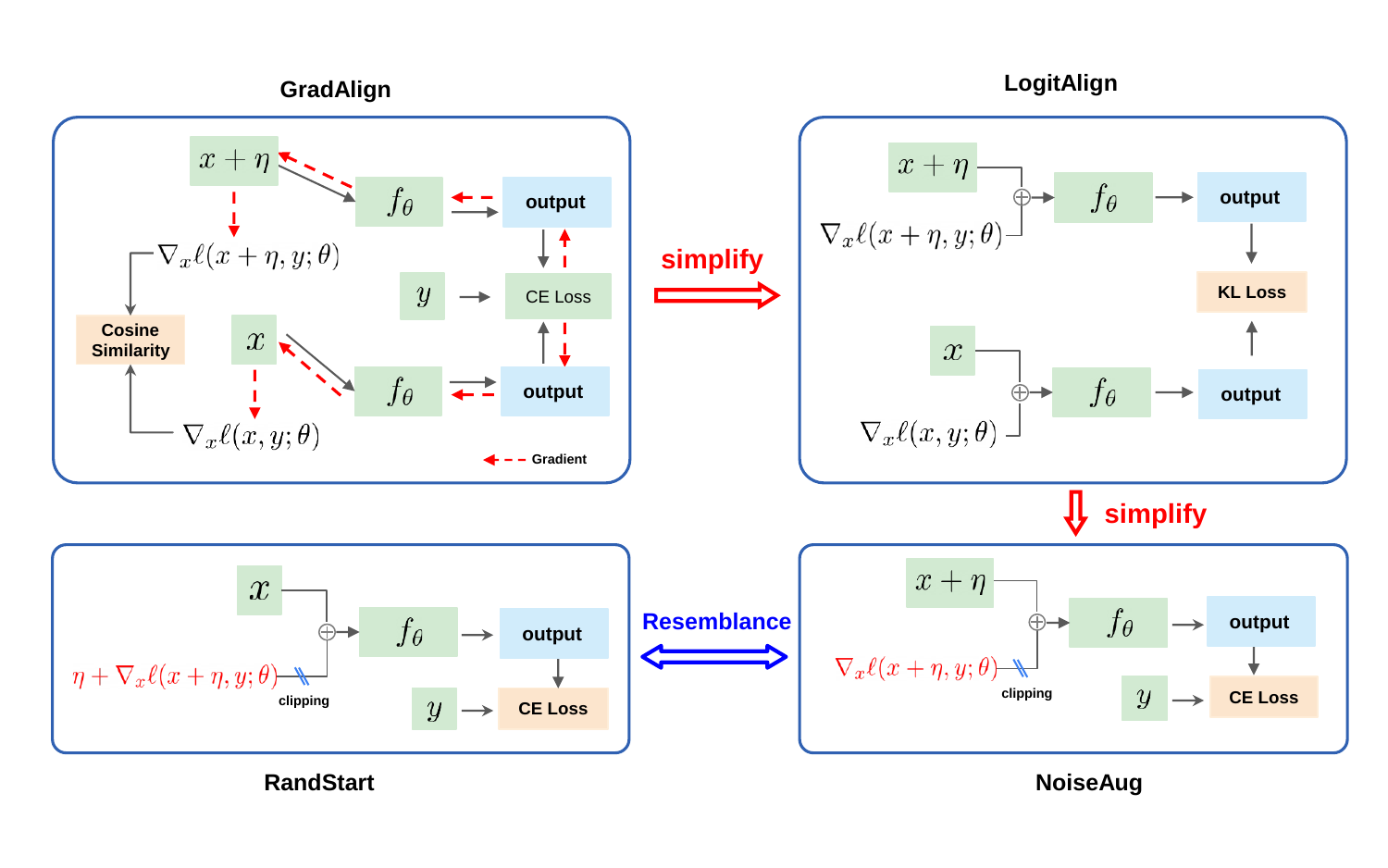}
\caption{Resemblance between RandStart and NoiseAug (a simplified variant GradAlign). The orange frame indicates the loss used to optimizing model parameters and
\textcolor{NavyBlue}{\textbf{\textbackslash\textbackslash}}
indicates perturbation clip between $-\epsilon$ and $\epsilon$. Original GradAlign encourages the similarity of gradients between clean example and its noisy counterpart ($\eta$ indicating the added random noise). LogitAlign simplifies GradAlign by aligning the logit output of adversarial examples generated on clean example and its noise counterpart. 
NoiseAug simplifies LogitAlign by replacing logit output of adversarial examples generated on clean example with ground-truth label. 
Simplified from GradAlign, NoiseAug has interesting resemblance with RandStart, both injecting random noise to the model input. 
The difference between RandStart and NoiseAug lies in that the added random noise is treated part of adversarial perturbation in RandStart and part of image in NoiseAug.
}\label{fig:GradAlign_simplication}
\end{center}
\end{figure*}

Simplified from GradAlign, our NoiseAug has surprising resemblance with RandStart (see Figure \ref{fig:GradAlign_simplication}), both boiling down to injecting noise to the model inputs (the difference between RandStart and NoiseAug is discussed in Sec. \ref{sec:Resemblance}). With such a high resemblance, the success of NoiseAug bridges the gap between the two popular approaches, \ie\ RandStart and GradAlign, in FGSM AT with a focus on the effect of noise injection on robustness performance. This helps interpret their success from a more unified perspective.
Such a unified perspective also yields an interesting question: why does a computation-free noise injection allow FGST AT with a larger perturbation size? Noise injection, with our NoiseAug as an exmple, can be perceived as an data augmentation method, thus we investigate whether this benefit is caused by data augmentation effect, \ie\ more training samples. We find that data augmentation effect is not its reason, instead, it can be attributed to the improved local linearity, which aligns with the finding in~\cite{andriushchenko2020understanding} that local non-linearity is the cause of CO. Moreover, we extend the local linearity in GradAlign~\cite{andriushchenko2020understanding} to a more generalized concept: model sensitivity to noise by measuring the influence of noise on the mode response. This insight helps explain why the GradAlign's simplified variants, including LogitAlign and NoiseAug, improve the local linearity. 

Overall, the contributions of our work are summarized as follows:

\begin{itemize}

    \item We perform an investigation of simplifying GradALign, yielding a new variant that has a high resemblance with simple RandStart. Such a high resemblance provides a more unified perspective on their success with the focus on injecting noise to the inputs.
    
    \item Such a unified perspective motivates to understand how noise injection improves the quality of FGSM under a large perturbation size. We verify that this benefit is not caused by data augmentation effect but by improved local linearity. Moreover, we extend local linearity for explaining the cause of catastrophic overfitting in FGSM AT to a more generalized form.
    
    \item Extensive results confirm that, without causing additional computation overhead, our proposed NoiseAug achieves a new SOTA performance for FAST AT.

\end{itemize}

\section{Background and experimental setup}
\textbf{Standard \vs\ Adversarial Training.} Let's assume $\mathcal{D}$ is a data distribution with $(x, y)$ pairs and $f(\cdot ,  \theta)$ is a model parameterized by $\theta$. Standard training (ST) minimizes the risk of $ \mathbb{E}_{(x,y) \sim \mathcal{D}} [l(f(x,\theta), y)]$, where $l$ indicates the cross-entropy loss. By contrast, adversarial training (AT) finds model parameter $\theta$ to optimize an adversarial risk:
\begin{equation}
\underbrace{\arg\min_{\mathbf{\theta}} \mathbb{E}_{(x,y) \sim \mathcal{D}} \underbrace{\left[ \max_{\delta \in \mathbb{S}} l(f(x + \delta; \theta), y) \right]}_{\mbox{inner maximization}}}_{\mbox{outer minimization}},
\label{eq:madry}
\end{equation}
where $\mathbb{S}$ denotes the allowed perturbation budget that is a typically $l_p$ norm-bounded $\epsilon$. A key difference between ST and AT is that AT generates adversarial examples as an inner maximization problem before optimizing the model weights. Following the convention in~\cite{andriushchenko2020understanding}, we mainly study AT under the constraint of $\epsilon = 8/255$ or $\epsilon = 16/255$.

\textbf{PGD \vs\ FGSM AT.} As discussed above, a unique nature of AT lies in solving an inner maximization problem. Projected gradient decent (PGD) is the most widely used approach for solving this problem. A critical hyperparameter in PGD-AT is the number of steps for generating adversarial examples. Following the convention in prior works~\cite{wong2020fast,andriushchenko2020understanding}, we term it PGD-$N$ when N steps are used. 
FGSM can be seen as a special case of PGD-$N$ when N is set to 1 and we 
term it FGSM AT. A major advantage of applying FGSM is that it makes the slow inner maximization more computation efficient. 

\textbf{Robust \vs\ Catastrophic Overfitting.} AT is widely known to suffer from overfitting. Specifically, PGD AT is found to have a decreasing robustness on the evaluation dataset in the later stage of training, which is termed robust overfitting (RO)~\cite{rice2020overfitting}. In general, RO is not a concern for FGSM AT with a shorter training schedule~\cite{wong2020fast}. However, FGSM AT is often subject to another more serious overfitting variant termed Catastrophic overfitting (CO)~\cite{wong2020fast}. Since the model is only trained on adversarial examples generated by FGSM, it has a risk of being overfitted to FGSM while losing its robustness against stronger attacks like PGD. In the latter stage of FGSM AT, the robustness against PGD attack might suddenly drop to zero when CO occurs. RO in PGD-AT is widely believed to be caused by lack of sufficient training samples~\cite{rice2020overfitting}, while CO in FGSM AT is attributed to local non-linearity of the model~\cite{andriushchenko2020understanding}. Since this work focuses on FGSM AT, we are mainly interested in CO. With low local linearity, the quality of FGSM attack decreases and thus cannot accurately solve the inner maximization problem~\cite{andriushchenko2020understanding}, causing a failure mode of AT. To decrease the step size is expected to avoid CO by improving the quality of FGSM attack~\cite{kim2020understanding} due to the assumption of linear nature~\cite{goodfellow2014explaining}. However, this leads to a sub-optimal solution against PGD attack with a larger perturbation budget.
In the following, we detail two popular strategies that avoid CO in FGSM AT with a larger step size.

\textbf{RandStart \vs\ GradAlign.} FGSM AT was long dismissed as ineffective against PGD attack~\cite{tramer2017ensemble}, however,~\cite{wong2020fast} shows that simple random start (RandStart), \ie\ initializing the perturbation with random noise, is sufficient for allowing successful FGSM AT with a larger step size like $\epsilon=8/255$. Other works~\cite{vivek2020single,kim2020understanding} also limit their investigation for the setup of $\epsilon=8/255$. Moreover, they either achieves inferior performance~\cite{vivek2020single} or (partly) relies on PGD attack~\cite{kim2020understanding}. Thus, this work mainly considers RandStart as an important baseline for its simplicity and efficiency. Unfortunately, CO still occurs in FGSM AT with RandStart when the perturbation size is larger than $11/255$, which hinders its use for a larger $\epsilon$ ($\ell_\infty$ $16/255$ for instance). Since CO often occurs in the latter stage of fast AT, early stop can be a straightforward way to alleviate CO, however, inefficient training often leads to a sub-optimal robust model~\cite{andriushchenko2020understanding}. The core of their method is a regularization loss GradAlign which is termed as such due to its goal for increasing the gradient alignment, \ie\ a metric to measure local linearity of model. Despite its success, as acknowledged in~\cite{andriushchenko2020understanding}, a major drawback of GradAlign is that the alignment regularization on the gradient yields heavy computation overhead, which makes it 3 to 4 times slower than the counterpart with RandStart. Note that RandStart can be perceived as a computation-free operation.
\begin{equation}
    \mathbb{ E}_{(x,y)\sim D,\eta \sim{\large  \mathcal{U}}   ([-\varepsilon , \varepsilon]^{d} ) } [\cos (  \nabla _{x}\ell (x,y;\theta ), \nabla _{x}\ell (x+\eta ,y;\theta ))],
\label{eq:GradAlign}
\end{equation}

\textbf{Experimental setup.} Unless mentioned otherwise, we follow~\cite{wong2020fast,andriushchenko2020understanding} to train PreAct ResNet18~\cite{he2016identity} for 30 epochs with the cyclic learning rates~\cite{smith2017cyclical} and half-precision training~\cite{micikevicius2017mixed}. The maximal learning rate in the cyclic schedule is 0.3. For the perturbation budget, we adopt $\ell_\infty$-norm constraint and set $\epsilon$ to $8/255$ or $16/255$. Following prior works, we evaluate the adversarial robustness with the attack of PGD-50-10, \ie\ 50 iterations and 10 restarts, where the step size is set to $\alpha=\epsilon/4$. For the step size in the training, we follow the setup in~\cite{andriushchenko2020understanding} by setting $\alpha$ to $1.25\epsilon$ and $\epsilon/2$ for single-step (FGSM) and twp-step (PGD-2) AT, respectively. Note that the training is performed with half-precision for speeding up but the evaluation is always conducted with single-precision for fair comparison because limited numerical precision in the gradient calculation might overestimate the model robustness. All the experiments conduct on a NVIDIA TITAN RTX GPU.

\section{Towards simplifying GradAlign}

\textbf{Motivation.} Considering RandStart and GradAlign, there seems to be a trade-off between efficiency and effectiveness. It is desirable to have a new regularization method that is as effective as GradAlign for enabling FGSM AT with a large perturbation size but causes less, ideally zero, computation overhead. Without the practice of early stop, GradAlign~\cite{andriushchenko2020understanding} is \textit{the first yet so far the only one} found in the literature to report success of FGSM AT against PGD attack with $\epsilon=16/255$. Therefore, we seek a simplified variant of GradAlign for reducing its heavy computation overhead without sacrificing its merit of allowing a larger step size. In the following, we detail the simplication process which is summarized in Figure~\ref{fig:GradAlign_simplication}.

\subsection{Towards simplifying GradAlign}

\begin{figure}[ht]
\begin{center}
\includegraphics[width=1\columnwidth]{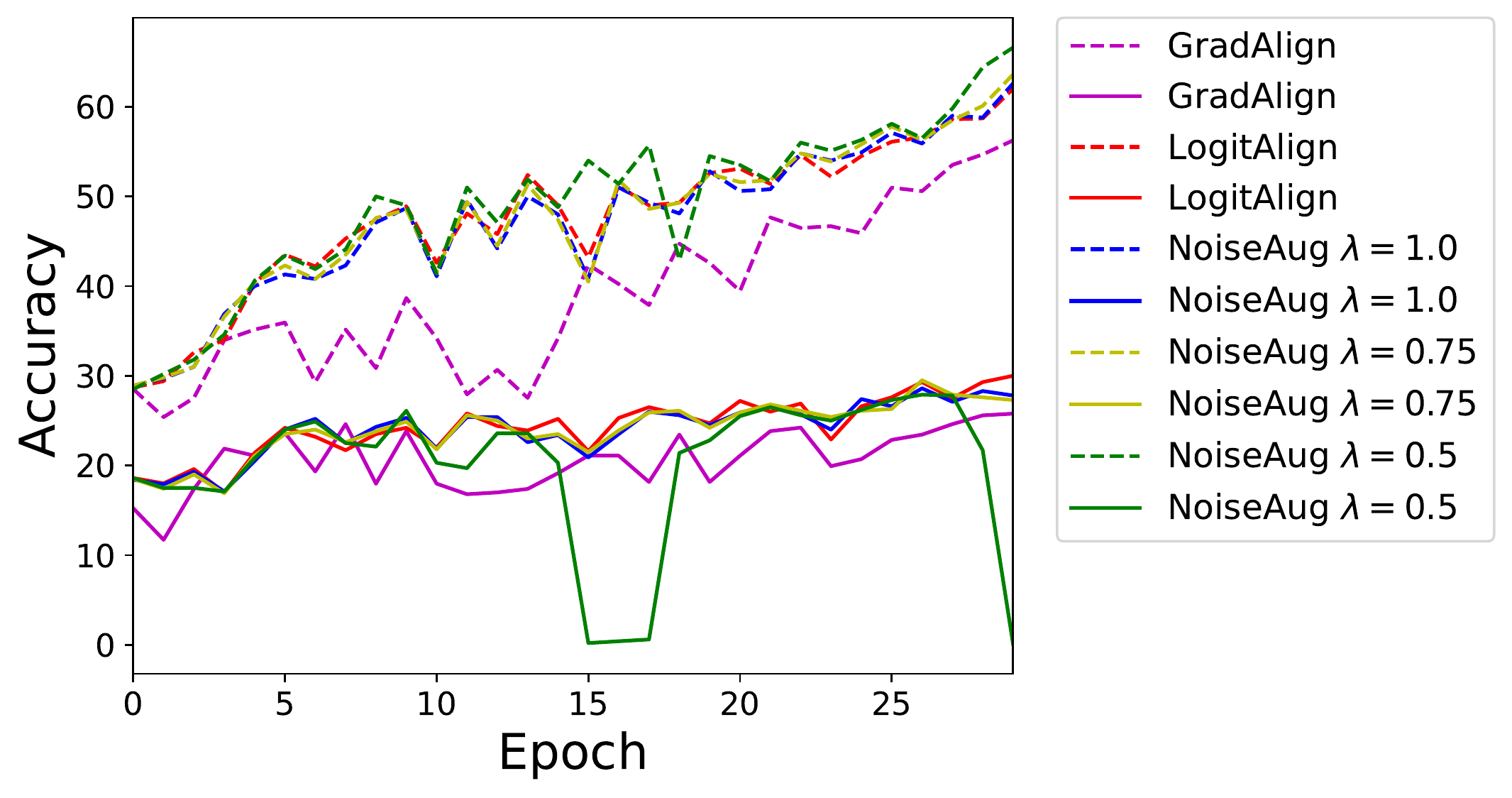}
\caption{Comparison among GradAlign, LogitAlign with KL regularization, and NoiseAug with various $\lambda$. Dash line indicates clean accuracy and solid line indicates accuracy under the attack of PGD-50-10.}
\label{fig:logitalign}
\end{center}
\end{figure}

\textbf{GradAlign to LogitAlign.} 
In essence, GradAlign encourages the model to behave similarly to clean example $x$ and corresponding noisy example $x+\eta$ ($\eta$ indicating random noise) for their input gradients. Given that such input gradient can be used for generating adversarial examples with FGSM, we conjecture that similar benefit might be obtained by directly aligning the logit output of the adversarial examples generated on clean examples and their noisy counterparts. Specifically, the total loss with a KL regularization to encourage such alignment is shown as:
\begin{equation}
    l(f(x + \delta_1; \theta),y) + KL(f(x + \delta_1; \theta),f(x+\eta + \delta_2; \theta)),
\label{eq:ce_combined_kl}
\end{equation}
where $\delta_1$ and $\delta_2$ are the adversarial perturbations for clean and noisy examples, respectively. Since the regularization is performed on the output logit, we term it LogitAlign to differentiate from GradAlign. An advantage of LogitAlign is that it avoids double backpropagation and thus facilitates the computation parallelization by simply concatenating the inputs in the PyTorch implementation. The results in Figure \ref{fig:logitalign} show that our proposed LogitAlign achieves performance superior to GradAlign but only requires around half the computation time. 
\textbf{LogitAlign to NoiseAug.}
As shown above, the alignment on both input gradient and output logit help avoid CO. Here, we further investigate whether alignment can be further removed. 

Following LogitAlign, we use adversarial examples from both clean examples and noisy examples in the outer minimization to update the model weights. By contrast, we do not use KL divergence to increase their similarity but optimize them directly with a combined CE loss as:
\begin{equation}
    (1-\lambda) l(f(x + \delta_1; \theta),y) + \lambda l(f(x+\eta + \delta_2; \theta),y).
\label{eq:ce_combined}
\end{equation}
where $\lambda \in [0, 1]$. As shown in Figure \ref{fig:logitalign}, we find that the increase of $\lambda$ helps alleviate CO and setting $\lambda$ to 1 achieves comparable performance as LogitAlign. Somewhat surprisingly, our empirical results suggest that the alignment is not absolutely necessary and simply augmenting images with noise might be sufficient for avoiding CO. Due to its simplicity as well as competitive performance, we adopt it as our final regularization method and term it \textit{NoiseAug}.

\begin{algorithm}[H]
\caption{NoiseAug-based FGSM AT for a network $f_\theta$ with $T$ epochs, given some radius $\epsilon$, adversarial step size $\alpha$ and $N$ PGD steps and a dataset with batch size $M$}
\label{alg:noise_aug}
\begin{algorithmic}
\FOR {$t=1\dots T$}
\FOR {$i=1\dots M$}
\STATE $x_i\longleftarrow x_i+ \eta$ \textit{// {Noise augmentation}}
\STATE $\delta=0$ \; 
\STATE $\delta = \delta + \alpha \cdot \textrm{sign}(\nabla_\delta \ell(f_\theta(x_i + \delta),y_i))$
\STATE $\delta = \max(\min(\delta, \epsilon), -\epsilon)$
\STATE $\theta = \theta - \nabla_\theta \ell(f_\theta(x_i + \delta), y_i)$ 
\ENDFOR
\ENDFOR
\end{algorithmic}
\end{algorithm}

\textbf{Noise augmentation is all you need for fast AT.} Compared with GradAlign and LogitAlign, a major advantage of NoiseAug is that it causes zero~\footnote{Technically, augmenting images with the noise still consumes computation, however, it is negligible compared with the time to generate adversarial examples and training the model.} additional overhead. Following the practice in~\cite{andriushchenko2020understanding}, our proposed NoiseAug by default initializes the perturbation as zero, rendering our single-step AT method as FGSM + NoiseAug. 
Without any conditional intervention like early stop, our single-step AT alleviates CO by combining the \textit{very first} FGSM AT method with a common noise augmentation. 
Implementation-wise, it is no more than a single line of code as shown in Algorithm~\ref{alg:noise_aug}.

\begin{table*}[]
\centering
\caption{Standard accuracy and robustness under the attack of PGD-50-10. Results with various augmentations.}
\label{tab:cross_test}
\resizebox{0.6\textwidth}{!}{  
\begin{tabular}{cccccc}
\toprule
\textbf{Augmentation} &\textbf{Standard} &  \textbf{PGD-50-10} & \textbf{FGSM} & \textbf{AA}  \\
\midrule
None        & 74.42$\pm$1.82  & 0.00$\pm$0.00  &  87.82$\pm$8.53   & 0.00$\pm$0.00  \\
Cutout~\cite{devries2017improved}   & 58.07$\pm$11.76 &	0.00$\pm$0.00  &  72.53$\pm$9.09   & 0.00$\pm$0.00 \\
Mixup~\cite{zhang2017mixup}           &  61.42$\pm$5.81 &	0.00$\pm$0.00  &  86.54$\pm$2.35   & 0.00$\pm$0.00 \\
CutMix~\cite{yun2019cutmix}          & 70.78$\pm$4.25  &	0.00$\pm$0.00  &  72.64$\pm$15.30  & 0.00$\pm$0.00 \\
NoiseAug (Ours)        & 59.18$\pm$0.37  & 29.40$\pm$0.33 &  36.44$\pm$0.63   & 21.24$\pm$0.38 \\
\bottomrule
\end{tabular}
}
\end{table*}

\subsection{Resemblance with RandStart and a unified perspective}\label{sec:Resemblance}

Simplified from GradAlign, our NoiseAug has a surprising resemblance with RandStart in the sense that both adopt random noise for improving FGSM AT. This contributes to understanding them with a unified perspective: input noise is the key for preventing CO. Despite such a unified perspective, there is a trivial-looking yet significant difference between our NoiseAug and existing RandStart.
\begin{table}[]
    \centering
    \caption{NoiseAug with random initialization.
    Random initialization leads to a higher standard accuracy but at the cost of a lower robustness.}
    \label{tab:zero_random_init}
    \resizebox{0.45\textwidth}{!}{ 
    \begin{tabular}{cccc}
    \toprule
     & \textbf{$\delta$ init}   & \textbf{Standard}  & \textbf{PGD-50-10} \\
    \midrule
    \multirow{2}{*}{$\epsilon=8/255.$} & zero    & 80.54$\pm$0.30 &  48.16$\pm$0.49    \\ 
    \cline{2-4}
     & random   & 82.38$\pm$0.02  &46.36$\pm$0.02     \\
     \midrule
    \multirow{2}{*}{$\epsilon=16/255.$} & zero   &  59.18$\pm$0.37    & 29.40$\pm$0.33    \\
    \cline{2-4}
    & random & 66.25$\pm$0.14  &25.84$\pm$0.04      \\
    \bottomrule
    \end{tabular}}
\end{table}

\textbf{Difference between NoiseAug and RandStart.} NoiseAug can be interpreted as separating the combined $\delta$ in RandStart into a disentangled random noise and FGSM perturbation. This disentangling might look trivial but yield a significant difference. First, due to constraint on perturbation magnitude, the initialized random noise in RandStart can not be set to be larger than $\mathcal{U}(-\epsilon, \epsilon)$. Our results in Table~\ref{tab:ablation_noise} suggest that an appropriate noise magnitude is critical for achieving high robustness. Disentangling random noise from the perturbation, NoiseAug allows to freely choose the optimal noise type or magnitude by just treating it as data augmentation. Second, random initialization can yield inferior performance (See Table \ref{tab:zero_random_init}). It is worth mentioning that~\cite{andriushchenko2020understanding} also shows that removing random initialization results in a stronger PGD-2 baseline. This motivates GradAlign and our NoiseAug to not use the random initialization as in RandStart. 

On the one hand, this result suggests that local linearity can be increased by directly regularizing on the input itself, which is overhead-free and renders the regularizing on input gradient like GradAlign unnecessary. On the other hand, it also challenges a claim in~\cite{andriushchenko2020understanding} regarding why random initialization in RandStart helps mitigate CO. Specifically, ~\cite{andriushchenko2020understanding} claims that it ``\textit{boils down to reducing the average magnitude of the perturbations}". Note that random noise in NoiseAug does not decrease the perturbation magnitude but it still helps avoid CO. 

\section{Why does noise injection help avoid CO in FGSM AT?}

\subsection{From preliminary guess to local linearity}

\textbf{Preliminary guess.} 
It is natural to ask why such a simple regularization is so effective to avoid CO. A preliminary guess is that  the benefit of NoiseAug in FGSM AT comes from data augmentation effect, \ie\ more training samples.

\begin{figure}[ht]
\begin{center}
\centerline{\includegraphics[width=0.9\columnwidth]{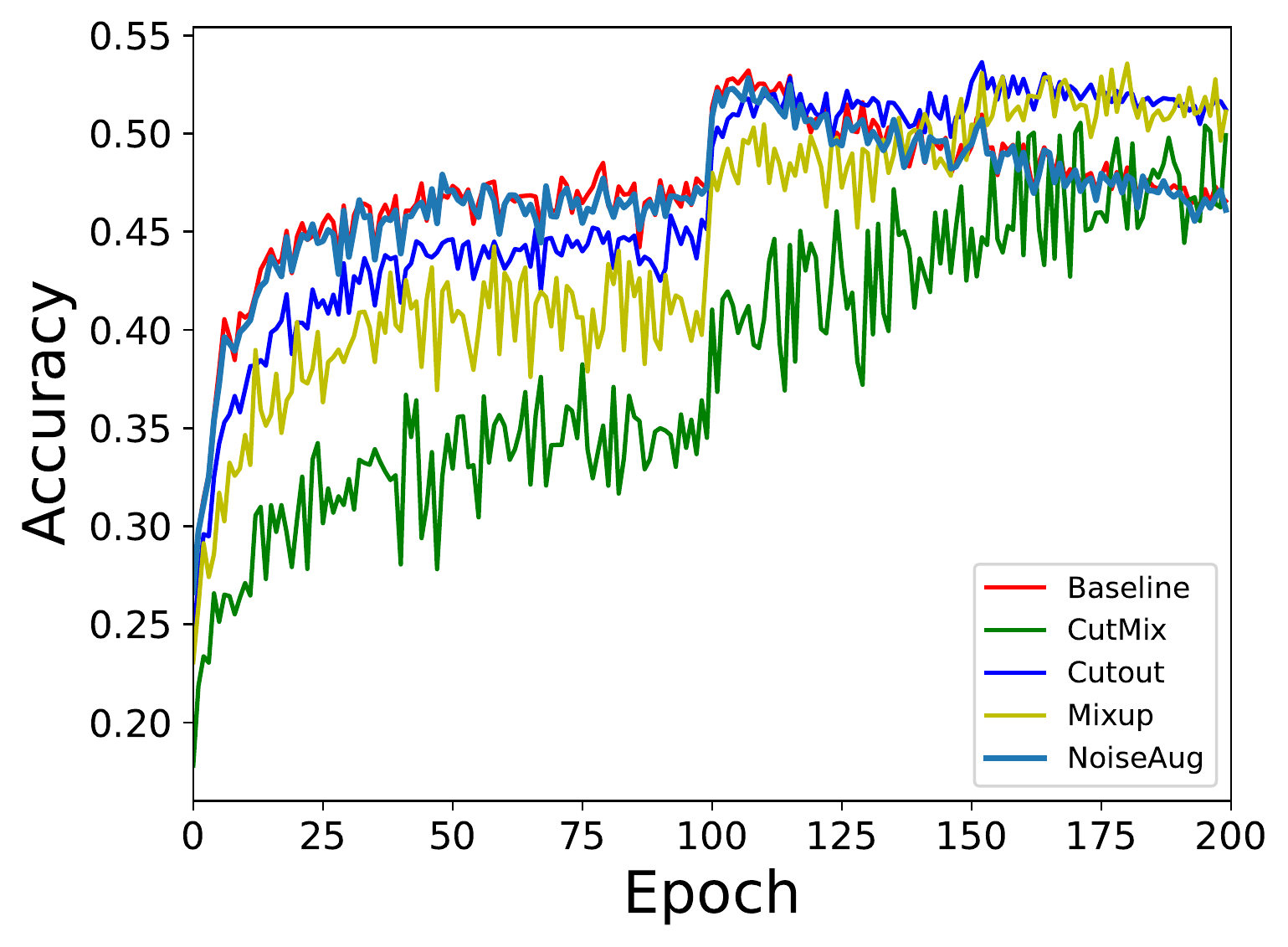}}
\caption{Robustness under the attack of PGD-10. CutMix, Coutout, and Mixup help alleviate robust overfitting (RO), while NoiseAug has little benefit for avoiding RO.}
\label{fig:cutoutmixup}
\end{center}
\end{figure}

Here, we experiment with other data augmentation methods, such as Mixup~\cite{zhang2017mixup}, Cutout~\cite{devries2017improved}, CutMix~\cite{yun2019cutmix}, which are known to be strong data augmentation methods for alleviating robust overfitting (RO)~\cite{rebuffi2021fixing}. The results in Table~\ref{tab:cross_test} show that none of the above methods helps alleviate CO except NoiseAug, suggesting the benefit of preventing CO does not come from data augmentation effect. This is further corroborated by the fact that NoiseAug cannot prevent RO as other augmentation methods. Following the setup in~\cite{rebuffi2021fixing}, we investigate the effectiveness of our NoiseAug to alleviate RO and the results are reported in Figure~\ref{fig:cutoutmixup}. CutMix, Cutout ,and Mixup mitigate the RO, which aligns the finding in~\cite{rebuffi2021fixing}. Moreover, unlabeled data is also found in~\cite{carmon2019unlabeled,gowal2021improving} to alleviate RO for improving robustness. The findings support that RO can be mitigated by creating more training samples. Interestingly, we find that noise-type augmentation brings little benefit to mitigate RO, suggesting Noise-type might not be as strong as other augmentations for generating more training samples. This is somewhat reasonable because the extra samples created by CutMix, Cutout ,and Mixup might be more diverse than just adding noise. It is worth mentioning that~\cite{andriushchenko2020understanding} also reports that, like our NoiseAug, their GradAlign does not prevent RO. This suggests that RO and CO might be less related as their shared term "overfitting" could imply. 
After rejecting this preliminary guess, in the following, we investigate its reason from an alternative perspective~\cite{andriushchenko2020understanding}.

\textbf{Local linearity and our conjecture.}
The occurrence of CO in FGSM AT can be seen as a result of FGSM attack failing to solve the inner maximization problem~\cite{andriushchenko2020understanding}. Taking the linear nature of FGSM attack into account, the authors of~\cite{andriushchenko2020understanding} argue that their GradAlign explicitly maximizes the gradient alignment for a higher local linearity, which improves the quality of FGSM solution and thus allows a larger step size without CO. Formally, the local linearity is defined in~\cite{andriushchenko2020understanding,benz2020batch} as follows:
\begin{equation}
    \mathbb{ E}_{(x,y)\sim D,\eta \sim{\large  \mathcal{U}}   ([-\varepsilon , \varepsilon]^{d} ) } [\cos (  \nabla _{x}\ell (x,y;\theta ), \nabla _{x}\ell (x+\eta ,y;\theta ))],
\label{eq:GradAlign}
\end{equation}
where $\eta$ is randomly sampled from a certain (like uniform) distribution. In this work, we follow the above definition for consistency and simplicity. The fact that GradAlign works by improving local linearity and that NoiseAug is simplified from GradAlign motivates our following conjecture.

\begin{table*}[!htbp]
    \centering
    \caption{FGSM AT method comparison on CIFAR10 dataset. Time is relative to the standard training.}
    \label{tab:method_comparision}
    \resizebox{0.7\textwidth}{!}{  
    \begin{tabular}{ccccccc}
    \toprule
    \textbf{Step}& \textbf{Method} &\textbf{Standard} &       \textbf{PGD-50-10} &  \textbf{FGSM} &  \textbf{AA} & \textbf{Time} \\
    \hline
    \textbf{none} & Standard & 94.04$\pm$0.19 &  0.00$\pm$0.00&  0.00$\pm$0.00&  0.00$\pm$0.00& 1 \\
    \hline
    \multicolumn{6}{c}{$\epsilon$=8/255}\\
    \hline
    \textbf{PGD-10} & Baseline & 81.64$\pm$0.64 &  50.62$\pm$0.42 &  55.76$\pm$1.01 & 47.26$\pm$0.53& $\sim$11 \\
    \hline
    \multirow{5}{*}{\textbf{FGSM}} & Baseline & 86.24$\pm$1.00& 0.02$\pm$0.04 &  86.66$\pm$9.17 &  0.00$\pm$0.00 & $\sim$2\\
    &RandStart~\cite{wong2020fast} & 84.50$\pm$0.44 &	45.34$\pm$0.45 &	54.04$\pm$0.36 &	42.86$\pm$0.14 & $\sim$2 \\
    &GradAlign~\cite{andriushchenko2020understanding} & 80.68$\pm$0.46 &  47.14$\pm$0.38 &  54.00$\pm$0.96 & 43.76$\pm$0.43 & $\sim$8 \\
    & NoiseAug (Ours) &80.54$\pm$0.30 &  48.16$\pm$0.49 &  54.52$\pm$0.48 & 44.48$\pm$0.44 & $\sim$2\\
    \hline
    \multicolumn{6}{c}{$\epsilon$=16/255}\\
    \hline
    \textbf{PGD-10} &   Baseline & 60.86$\pm$0.78 &  32.86$\pm$0.54 &  39.08$\pm$0.76 & 25.48$\pm$0.48 \\
    \hline
    \multirow{5}{*}{\textbf{FGSM}}&Baseline & 74.42$\pm$1.82 &   0.00$\pm$0.00 &  87.82$\pm$8.53 &  0.00$\pm$0.00 & $\sim$2\\
    &RandStart~\cite{wong2020fast} & 71.72$\pm$8.13 & 	0.00$\pm$0.00 & 	66.32$\pm$22.67 & 	0.00$\pm$0.00 & $\sim$2\\
    &Gradalign~\cite{andriushchenko2020understanding} & 59.06$\pm$0.75 &  27.00$\pm$0.44 &  34.60$\pm$0.39 & 18.56$\pm$0.34 & $\sim$8\\
    &NoiseAug (Ours) & 59.18$\pm$0.37 &  29.40$\pm$0.33 &  36.44$\pm$0.63 & 21.24$\pm$0.38 & $\sim$2\\
    \bottomrule
    \end{tabular}}
\end{table*}

\begin{figure}[ht]
\begin{center}
    \includegraphics[width=.9\linewidth]{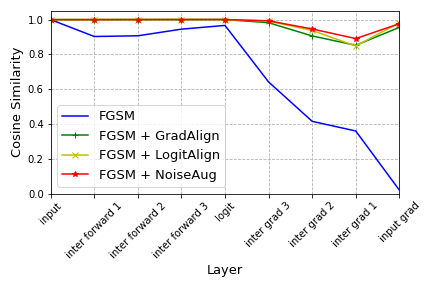} 
    \caption{Noise sensitivity of different method. We measure the cosine similarity between intermediate feature map of clean and noised image.}
\label{fig:noiseaug_sensitive}
\end{center}
\end{figure}

\begin{figure*}[!t]
\centering
\subfloat[]{\includegraphics[width=3in]{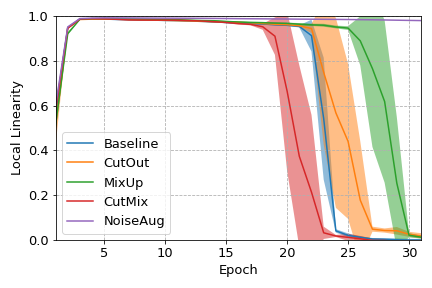}%
\label{fig_first_case}}
\hfil
\subfloat[]{\includegraphics[width=3in]{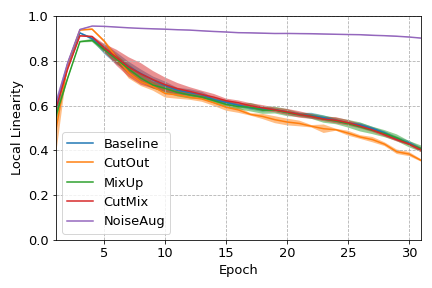}%
\label{fig_second_case}}
\caption{Visualization of the local linearity changing during training of FGSM+AT and standard training trained on ResNet-18 with CIFAR-10 under $\epsilon=8/255$.
Among the investigated four types of Augmentations, only NoiseAug helps increase local linearity, while CutMix, Cutout, Mixup have no effect in increasing local linearity. (a) FGSM+AT, (b) FGSM+AT. }
\label{fig:linearity_AT}
\end{figure*}

\textbf{Conjecture:} NoiseAug helps FGSM AT avoid CO because it increases local linearity.

Adopting the same metric in~\cite{benz2020batch,andriushchenko2020understanding}, we investigate whether augmenting images with noise improves local linearity. 
The results in Figure~\ref{fig:linearity_AT} (a) show that NoiseAug significantly improves the local linearity,
which verifies our above conjecture. By contrast, other data augmentation methods do not improve the local linearity.

\textbf{Local linearity in standard training.} Our above analysis shows that NoiseAug improves the local linearity in FGSM AT. We further investigate the influence of NoiseAug in standard training and the results are reported in Figure~\ref{fig:linearity_AT} (b). The local linearity decreases when the training epoch increases, which aligns with the finding in~\cite{benz2020batch,andriushchenko2020understanding}. Compared with the baseline, NoiseAug improves local linearity in the whole training process by a significant margin.

\subsection{Why does NoiseAug improve local linearity?}
\textbf{Explicit \vs\ implicit regularization.} GradAlign~\cite{andriushchenko2020understanding} adopts an explicit regularization loss to increase the local linearity, thus the improved local linearity in GradAlign is well expected. Without such an explicit regulatization loss, it is unclear why NoiseAug also fulfills the same purpose. Here, we attempt to provide an intuitive explanation based on a new interpretation of local linearity.

\textbf{From local linearity on input gradients to the noise sensitivity of model.} The input gradient in GradAlign is calculated through both forward propagation (from input to logit) and backward propagation (from logit to input gradient). The local linearity denoted as the cosine similarity of input gradients can be interpreted as \textit{a metric of how sensitive the whole process (forward propagation + backward propagation) is to random noise}. GradAlign~\cite{andriushchenko2020understanding} only evaluates the final input gradient similarity, however, more insight can be derived by checking the similarity for intermediate features and their gradients. The results in Figure~\ref{fig:noiseaug_sensitive} show that along the whole forward and backward propagation, GradAlign and our NoiseAug consistently increase the cosine similarity, \ie\ reducing the sensitivity to noise. Here, we provide a more general interpretation of local linearity as how sensitive the model is to random noise. Local linearity can be seen as a special case of our interpretation by only observing the input gradient similarity.

\begin{figure*}[ht]
\centering
\subfloat[]{\includegraphics[width=2 in]{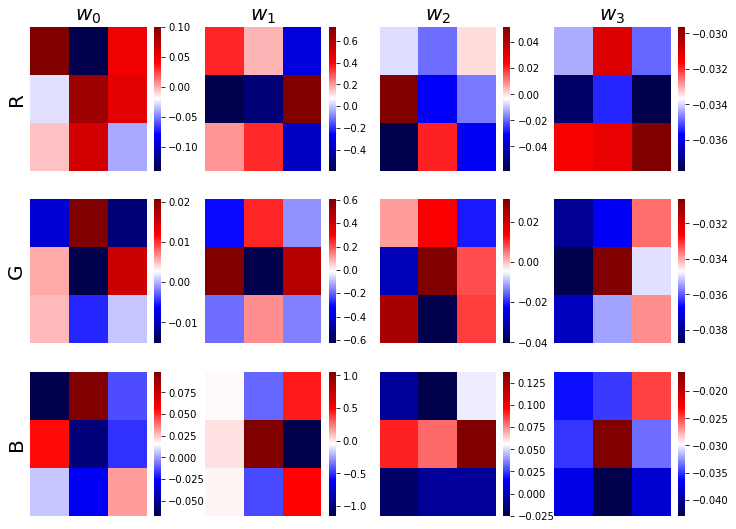}%
\label{fig_first_case}}
\hfil
\subfloat[]{\includegraphics[width=2 in]{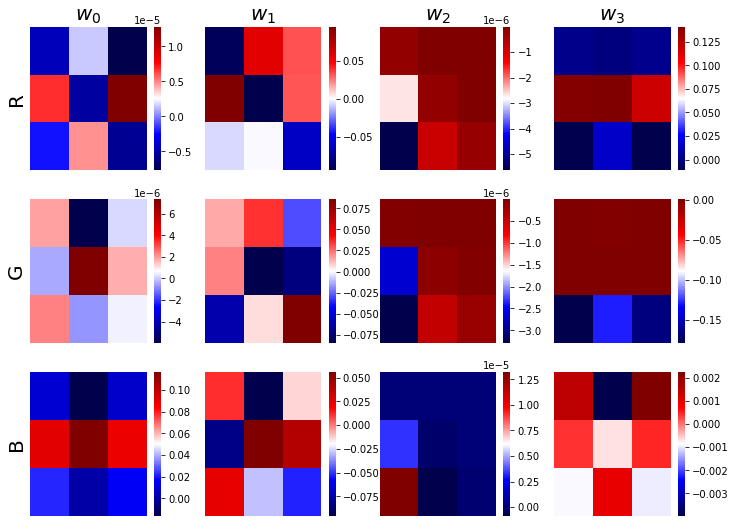}%
\label{fig_second_case}}
\hfil
\subfloat[]{\includegraphics[width=2 in]{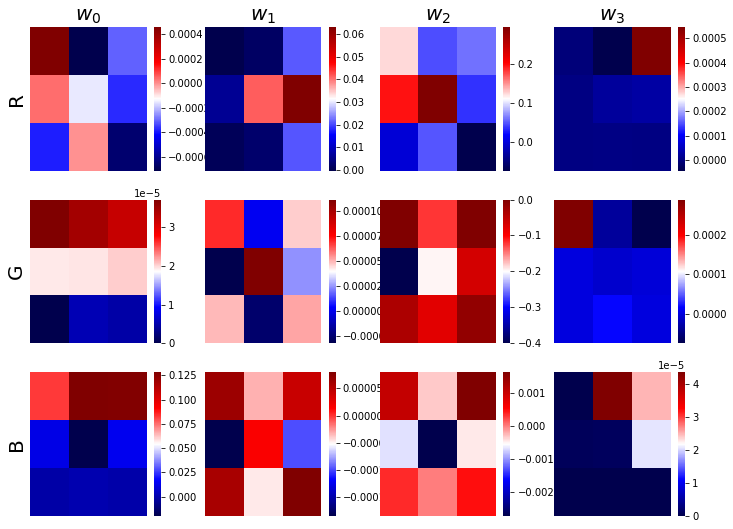}%
\label{fig_second_case}}
\caption{Filter Visualization in FGSM AT, GradAlign, and NoiseAug. We can only observe a Laplace filter in the model trained with FGSM AT, G (green) channel of $w_1$. }
\label{fig:baseline_filter}
\end{figure*}

\begin{figure*}[ht]
\centering
\begin{center}
\centerline{\includegraphics[width= 12cm]
{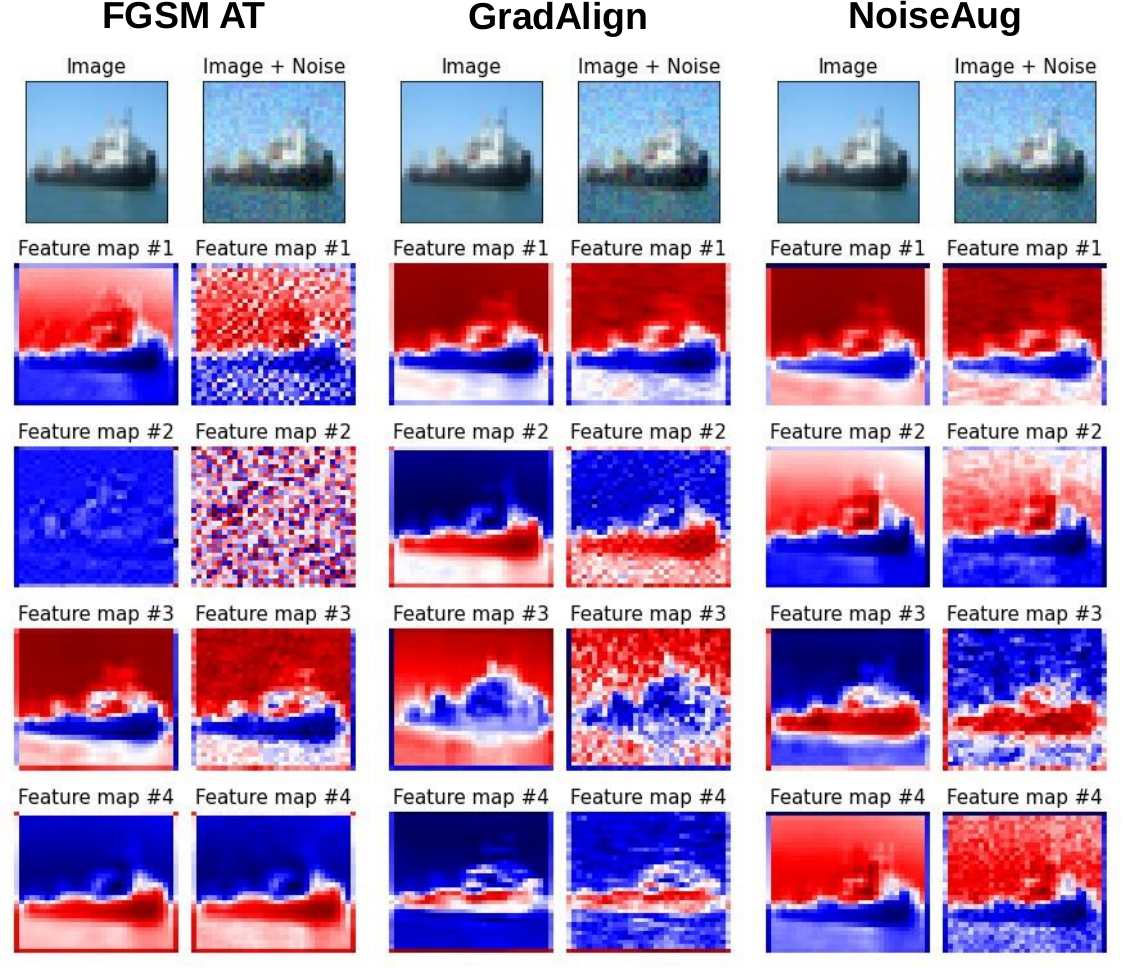}} 
\caption{Feature map comparison on different methods, \ie~FGSM AT, GradAlign, and NoiseAug. Here we plot the intermediate feature map processing with the learned weights on a 4 layer CNN.}
\label{fig:img4_feature_map}
\end{center}
\end{figure*}

\textbf{Feature layer analysis on a toy example.} To further investigate the sensitivity of model with noise input, we conduct a toy experiment on a small network and compare both the trained model weights and intermediate feature map between different methods. Specifically, following~\cite{andriushchenko2020understanding}, we train a single-layer CNN with 4 filters on CIFAR10 using $\epsilon=10/255$ with 30 epochs. The results are shown in Figure~\ref{fig:baseline_filter}. Our results confirm that vanilla FGSM AT has a noise sensitive Laplace filter (see Green channel in $w_1$) and GradAlign has no such filters. Our NoiseAug also has no such filters. Figure~\ref{fig:img4_feature_map} shows that there is a feature map very sensitive to noise in vanilla FGSM AT (caused by the Laplace filter) but not in GradALign and NoiseAug.

\begin{table*}[!htbp]
\centering
\caption{PGD2 method comparison on CIFAR10.}
\label{tab:pgd2}
\resizebox{0.7\textwidth}{!}{  
\begin{tabular}{ccccccc}
\toprule
\textbf{Step}& \textbf{Method} &\textbf{Standard} &       \textbf{PGD-50-10} &  \textbf{FGSM} &  \textbf{AA} \\
\hline
\multicolumn{6}{c}{$\epsilon$=16/255}\\
\hline
\textbf{PGD-10} &   Baseline & 60.86$\pm$0.78 &  32.86$\pm$0.54 &  39.08$\pm$0.76 & 25.48$\pm$0.48 \\
\hline
\multirow{3}{*}{\textbf{PGD-2}}&   baseline & 62.54$\pm$3.23 & 11.78$\pm$13.04 & 48.04$\pm$10.23 &  7.82$\pm$9.32 \\
 & Gradalign~\cite{andriushchenko2020understanding} & 59.48$\pm$0.46 &  30.30$\pm$0.44 &  36.12$\pm$0.68 & 21.96$\pm$0.27 \\
 \cline{2-6}
 &  NoiseAug (Ours) & 59.08$\pm$0.48 &  30.72$\pm$0.55 &  35.54$\pm$0.48 & \textbf{23.02$\pm$0.51} \\
\bottomrule
\end{tabular}
}
\end{table*}

\textbf{Input noise augmentation as the critical component.} With the above interpretation, NoiseAug would be the most straightforward method to decrease the model sensitivity to noise. GradAlign and LogitAlign also have noise on the input, which naturally explain why they also decrease the model sensitivity as NoiseAug. Since the key element for reducing noise sensitivity boils down to input noise augmentation, any additional components that cause additional computation overhead become redundant. Therefore, the key reason for the success of NoiseAug lies in keeping the critical component for reducing noise sensitivity but removing the redundant components in GradAlign.

\begin{table*}[h]
\centering
\caption{Different method comparison on SVHN dataset. With epsilon 8/255 and 12/255.}
\label{tab:result_svhn}
\resizebox{0.7\textwidth}{!}{ 
\begin{tabular}{cccccc}
\toprule
 \textbf{Step} &\textbf{Method} &\textbf{Standard} &\textbf{PGD-50-10} &\textbf{FGSM}  &\textbf{AA} \\
\midrule
\multicolumn{6}{c}{$\epsilon=8/255$}\\
\hline
\multirow{1}{*}{\textbf{PGD-10}} &   Baseline & 83.82$\pm$0.79 & 38.28$\pm$0.44 &  48.92$\pm$0.41 & 30.36$\pm$0.53 \\
\hline
\multirow{5}{*}{\textbf{FGSM}} &   Baseline & 90.58$\pm$0.57 &  0.00$\pm$0.00 &  93.56$\pm$4.16 &  0.00$\pm$0.00 \\
 &   RandStart~\cite{wong2020fast} &95.62$\pm$0.72 &	0.36$\pm$0.44 &	84.78$\pm$9.26 &	0.00$\pm$0.00 \\
 & GradAlign~\cite{andriushchenko2020understanding} & 91.74$\pm$0.66 & 42.52$\pm$0.69 &  59.66$\pm$0.79 & 39.52$\pm$0.53 \\
\cline{2-6}
 &  NoiseAug (Ours) & 91.00$\pm$0.21 & 44.54$\pm$0.62 &  58.50$\pm$0.58 & \textbf{41.54$\pm$0.36} \\
  \bottomrule
  \toprule
\multicolumn{6}{c}{$\epsilon=12/255$}\\
\hline
\multirow{1}{*}{\textbf{PGD-10}} &   Baseline & 83.82$\pm$0.79 & 38.28$\pm$0.44 &  48.92$\pm$0.41 & 30.36$\pm$0.53 \\
\hline
\multirow{5}{*}{\textbf{FGSM}} &   Baseline & 87.58$\pm$2.30 &  0.00$\pm$0.00 & 86.64$\pm$11.01 &  0.00$\pm$0.00 \\
 &   RandStart~\cite{wong2020fast} & 94.44$\pm$0.69&	0.00$\pm$0.00&	68.00$\pm$17.32&	0.00$\pm$0.00 \\
 & GradAlign~\cite{andriushchenko2020understanding} & 84.32$\pm$0.68 & 24.06$\pm$0.22 &  43.06$\pm$0.30 & 19.64$\pm$0.67 \\
\cline{2-6}
 &  NoiseAug (Ours) & 84.58$\pm$0.43 & 27.92$\pm$0.56 &  46.68$\pm$0.71 & \textbf{23.04$\pm$0.53} \\
\bottomrule
\end{tabular}
}
\end{table*}

\begin{table*}
\centering
\caption{Different method comparison on WideResNet28-10 with CIFAR10 dataset $\epsilon=8/255$. results of YOPO, Free-AT, FGSM-CKPT, and SLAT come from Reliably Fast AT. For the required training time, we report
the value that is relative to the standard training.}
\label{tab:result_WideResNet28-10}
\resizebox{0.6\textwidth}{!}{ 
\begin{tabular}{cccccc}
\toprule
\textbf{Method} &     \textbf{Standard} &  \textbf{PGD-50-10} &  \textbf{AA} & \textbf{Time}\\
\midrule
PGD-10  & 84.82$\pm$0.35 & 54.58$\pm$0.81  & 51.72$\pm$0.50 & $\sim$11 \\
\hline
YOPO-5-3~\cite{zhang2019you}& 82.35$\pm$ 1.78&  34.23$\pm$ 3.61&   32.79$\pm$ 3.65& $\sim$1\\
Free-AT($m=8$)~\cite{shafahi2019adversarial}& 76.57$\pm$ 0.19&  44.15$\pm$ 0.30&  41.02$\pm$ 0.20& $\sim$2.1\\
FGSM  & 83.56$\pm$1.01 &  0.00$\pm$0.00  &  0.00$\pm$0.00 & $\sim$2\\
RandStart~\cite{wong2020fast} & 87.72$\pm$2.28 &  0.00$\pm$0.00  &  0.00$\pm$0.00& $\sim$2 \\
GradAlign~\cite{andriushchenko2020understanding} & 83.80$\pm$0.25 & 48.95$\pm$0.53  & 46.15$\pm$0.30& $\sim$8 \\
CKPT($c=3$)~\cite{kim2020understanding} & 89.32$\pm$ 0.10&  40.83$\pm$ 0.36&   39.38$\pm$ 0.24& $\sim$2.2 \\
SLAT~\cite{park2021reliably} & 85.91$\pm$ 0.31&  47.06$\pm$ 0.03&  44.62$\pm$ 0.11 & $\sim$2\\
\textbf{NoiseAug (Ours)} & 83.92$\pm$0.51 & \textbf{49.64$\pm$0.64}  & \textbf{47.24$\pm$0.27} & $\sim$2\\
\bottomrule
\end{tabular}
}
\end{table*}

\begin{table*}[]
\centering
\caption{NoiseAug with uniform distribution noise $\mathcal{U}$ and normal distribution noise $\mathcal{N}$ with magnitude in $\{ 1.0\times\epsilon, 2.0\times\epsilon, 3.0\times\epsilon, 4.0\times\epsilon,\}$ trained on CIFAR10 with PreAct ResNet18. The bold result indicate the most robust one under AutoAttack.}
    \label{tab:ablation_noise}
\resizebox{0.6\textwidth}{!}{ 
\begin{tabular}{c|c|ccccc}
\toprule
 \textbf{Type} & \textbf{Scale} & \textbf{Standard} & \textbf{PGD-50-10} &       \textbf{FGSM} &  \textbf{AA} \\
\midrule
\multicolumn{6}{c}{$\epsilon=8/255$}\\
\hline
      \multirow{4}{*}{$\mathcal{U}$} &        1.0$\times\epsilon$ & 81.82$\pm$0.57 &  47.10$\pm$0.55 &  54.66$\pm$0.59 & 44.60$\pm$0.61 \\
         &        2.0$\times\epsilon$ & 81.30$\pm$0.33 &  47.74$\pm$0.44 &  54.62$\pm$0.59 & 44.26$\pm$0.51 \\
         &        3.0$\times\epsilon$ & 80.54$\pm$0.30 &  48.16$\pm$0.49 &  54.52$\pm$0.48 & \textbf{44.48$\pm$0.44} \\
         &        4.0$\times\epsilon$ & 78.76$\pm$0.56 &  47.00$\pm$0.44 &  53.00$\pm$0.78 & 43.46$\pm$0.66 \\
   \hline
         \multirow{4}{*}{$\mathcal{N}$} &        1.0$\times\epsilon$ & 81.70$\pm$0.83 &  47.98$\pm$0.62 &  54.76$\pm$0.61 & 44.50$\pm$0.30 \\
          &        2.0$\times\epsilon$ & 80.18$\pm$0.20 &  48.06$\pm$0.54 &  54.38$\pm$0.77 & 44.00$\pm$0.57 \\
          &        3.0$\times\epsilon$ & 77.74$\pm$0.56 &  46.94$\pm$0.78 &  52.20$\pm$0.85 & 43.18$\pm$0.67 \\
          &        4.0$\times\epsilon$ & 75.62$\pm$0.51 &  45.50$\pm$0.82 &  50.26$\pm$0.70 & 41.36$\pm$0.71 \\
   \bottomrule
   \toprule
\multicolumn{6}{c}{$\epsilon=16/255$}\\
\hline
\multirow{4}{*}{$\mathcal{U}$} &        1.0$\times\epsilon$ & 61.02$\pm$3.69 & 16.36$\pm$13.39 & 44.12$\pm$25.89 & 11.48$\pm$9.41 \\
&        2.0$\times\epsilon$ & 62.10$\pm$0.45 &  28.30$\pm$0.21 &  37.54$\pm$0.40 & 20.46$\pm$0.68 \\
   &        3.0$\times\epsilon$ & 59.18$\pm$0.37 &  29.40$\pm$0.33 &  36.44$\pm$0.63 & \textbf{21.24$\pm$0.38} \\
   &        4.0$\times\epsilon$ & 57.94$\pm$0.48 &  28.56$\pm$0.55 &  34.06$\pm$0.33 & 20.94$\pm$0.30 \\
  \hline
         \multirow{4}{*}{$\mathcal{N}$} & 1.0$\times\epsilon$ & 62.44$\pm$0.58 &  28.74$\pm$0.66 &  38.48$\pm$0.60 & 20.30$\pm$0.94 \\
    &        2.0$\times\epsilon$ & 59.30$\pm$0.51 &  29.62$\pm$0.46 &  36.58$\pm$0.40 & 20.66$\pm$0.49 \\
    &        3.0$\times\epsilon$ & 56.32$\pm$0.69 &  28.96$\pm$0.62 &  34.12$\pm$0.50 & 20.24$\pm$0.65 \\
    &        4.0$\times\epsilon$ & 53.08$\pm$0.36 &  27.28$\pm$0.29 &  31.56$\pm$0.26 & 20.04$\pm$0.48 \\
\bottomrule
\end{tabular}
}
\end{table*}

\section{Experiment results}

\textbf{Main results.} FGSM AT + NoiseAug, here, adopts uniform distribution noise $\mathcal{U}\in[-3\epsilon,3\epsilon]$. Under the same setup, we compare the following methods: standard FGSM, RandStart, FGSM AT + GradAlign~\cite{andriushchenko2020understanding}. Each model is trained with 5 random seeds. The results on CIFAR 10 with PreAct ResNet18 are reported in Table~\ref{tab:method_comparision}. All the results are either retrieved from the original works or reproduced with their official code. Since~\cite{andriushchenko2020understanding} only reports results on a subset of the evaluation dataset, we reproduce it on the full evaluation dataset. The parameter $\lambda$ is set to 0.2 and 2 for $\epsilon=8/255$ and $\epsilon=16/255$, respectively, as their paper suggested. Among all single-step FGSM AT methods, our simple FGSM+NoiseAug performs the best. We highlight that 
GradAlign induce extra computation overhead to different degrees over the baseline FGSM AT, while our NoiseAug is overhead-free. RandStart is also overhead-free but it yields less satisfactory performance under $\epsilon=8/255$ and totally fails under $\epsilon=16/255$. For the more challenging setup with $\epsilon=16/255$, we also report the reuslts of PGD-2 in the Table~\ref{tab:pgd2}, which shows show that our NoiseAug also outperforms other methods with PGD-2 baseline, which further bridges its gap with PGD-10. In additation, we report the results on SVHN as shown in Table~\ref{tab:result_svhn}. We conduct this experiment following the setup in~\cite{andriushchenko2020understanding}, we train a PreAct ResNet18 using $\epsilon=8/255$ and $\epsilon=12/255$ on SVHN for 15 epochs. The results in Table~\ref{tab:result_svhn} show that our methods outperform existing methods by a non-trivial margin, which mirrors the trend on CIFAR10.

\textbf{Results on WideResNet28-10.} We further compare our NoiseAug with other methods on  WideResNet28-10~\cite{zagoruyko2016wide}. Specially, the involved methods include YOPO~\cite{zhang2019you}, AT for free~\cite{shafahi2019adversarial} and FGSM+CKPT~\cite{kim2020understanding}, SLAT~\cite{park2021reliably}. We retrieve the results from~\cite{park2021reliably}. 
As shown in Table~\ref{tab:result_WideResNet28-10}, our NoiseAug still achives the best robustness for both PGD attack and Auto attack.

\textbf{Noise type and scale.} Table~\ref{tab:ablation_noise} reports the influence of noise type and scale. For uniform noise, the basic magnitude is set to $\mathcal{U}(-\epsilon, \epsilon)$ and $\epsilon\times \mathcal{N}(0, 1)$ for Gaussian noise and uniform noise, respectively. Both of them are multiplied by a scale factor ($s$). We investigate the scale in the range from 0 to 3. The results show that both noise types significantly improve the robustness. Moreover, the standard accuracy decreases when the noise magnitude is set to too large.

\section{Related work}\label{sec:related}
It has been an active topic in the adversarial machine learning community to make AT more computation efficient. Here, we summarize the main progress in the past few years.

\textbf{``Free" to FGSM AT.} To make AT more computation efficient, there are two lines of works. Early attempts~\cite{shafahi2019adversarial,zhang2019you} investigated the possibility of ``free" adversarial training to achieve robustness with similar computation overhead as standard training. Another line of work attempt to minimize the number of steps to generate the adversarial examples. A major advantage of the second line of approaches is that it has extremely few parameters to tune, which makes it easily compatible with most training procedures. For example, it can be drastically accelerated~\cite{wong2020fast} by using standard techniques for boost training, such as cyclic learning rates~\cite{smith2018super} and mixed-precision training~\cite{micikevicius2017mixed}. In practice, ``Free" AT is not really free because their minibatch replays often require much more training steps even though it is faster than FGSM-AT for a single training step. Recently, the trend has shifted from multi-step ``free" AT to single-step FGSM AT. 

\textbf{Development of FGSM AT.} Since the first success of RandStart~\cite{wong2020fast} to show reasonable robustness against PGD-50-10, multiple works have attempted to improve FGSM AT with various techniques.~\cite{li2020towards} claims that the success of RandStart lies in improved success factor to recover from CO and proposed to use PGD when the CO is detected.~\cite{vivek2020single}~has claimed that the CO is caused by model parameter overfitting in the early stage, which motivates their dynamic dropout scheduling.~\cite{vivek2020single} has proposed to regularize the FGSM AT by introducing dropout layer after each non-linear layer. Specifically, those dropout layers are initialized with a high dropout probability which is linearly decayed during the training.~\cite{kim2020understanding} assumed that the overfitting is caused by a fixed perturbation magnitude and thus proposed to search a sample-wise minimum perturbation to avoid CO. 
Motivated from an observation that CO often occurs when the local linearity of the model is low, ~\cite{andriushchenko2020understanding} has introduced a regularization loss (GradAlign) to explicitly maximize local linearity for avoiding CO.

\section{Closing remark}
This work has studied how to improve FGSM AT, especially in terms of preventing CO. With the motivation to avoid double backpropagation in GradAlign, our investigation shows that LogitAlign achieves comparable performance. More interestingly, we find that simply augmenting images with noise achieves the best performance. Despite the simplicity, our proposed NoiseAugment outperforms existing regularization methods by a visible margin yet causes zero computation overhead. We investigate why NoiseAugment improves FGSM AT. Specifically, we have performed a comprehensive study on CO and RO through the lens of data augmentation and found that they need different augmentations. We have shown that in both AT and standard training, only Noise-type augmentation improves local linearity of model and thus improve FGSM AT.

\bibliographystyle{IEEEtran}
\bibliography{bib_mixed}

\vfill

\end{document}